\PassOptionsToPackage{table}{xcolor} 
\documentclass[runningheads]{llncs}

\usepackage{eccv}

\usepackage{eccvabbrv}
\usepackage{tikz}
\usepackage{pifont}

\usepackage{graphicx}
\usepackage{booktabs}
\usepackage{makecell}
\usepackage{multirow}
\usepackage{wrapfig}
\usepackage{arydshln}
\usepackage{algorithm}
\usepackage{algpseudocode}
\usepackage{placeins} 
\usepackage{float}

\usepackage[table]{xcolor} 
\usepackage{colortbl}

\newcommand{\tabincell}[2]{\begin{tabular}{@{}#1@{}}#2\end{tabular}}

\definecolor{cvprblue}{rgb}{0.21,0.49,0.74}

\usepackage[accsupp]{axessibility}  

\usepackage[breaklinks,colorlinks,citecolor=eccvblue]{hyperref}

\usepackage{orcidlink}
\makeatletter
\renewcommand\paragraph{\@startsection{paragraph}{4}{\z@}%
  {-1ex \@plus -0.5ex \@minus -0.2ex}%
  {-1em}%
  {\normalfont\bfseries}}
\makeatother

\usepackage{graphicx}      
\usepackage{amssymb}       
\usepackage[table]{xcolor} 
\usepackage{multirow}
\usepackage{float}
\begin{document}

\title{DiveUp: Learning Feature Upsampling from Diverse Vision Foundation Models} 

\titlerunning{DiveUp}

\author{Xiaoqiong Liu\inst{1} \and Heng Fan\inst{1}}

\authorrunning{X.~Liu \and H.~Fan} 

\institute{Department of Computer Science and Engineering, University of North Texas\\
\email{xiaoqiongliu@my.unt.edu, heng.fan@unt.edu}}

\maketitle

\begin{abstract}

Recently, feature upsampling has gained increasing attention owing to its effectiveness in enhancing vision foundation models (VFMs) for pixel-level understanding tasks. Existing methods typically rely on high-resolution features from the same foundation model to achieve upsampling via self-reconstruction. However, relying solely on intra-model features forces the upsampler to overfit to the source model's inherent location misalignment and high-norm artifacts. To address this fundamental limitation, we propose DiveUp, a novel framework that breaks away from single-model dependency by introducing multi-VFM relational guidance. Instead of naive feature fusion, DiveUp leverages diverse VFMs as a panel of experts, utilizing their structural consensus to regularize the upsampler's learning process, effectively preventing the propagation of inaccurate spatial structures from the source model. To reconcile the unaligned feature spaces across different VFMs, we propose a universal relational feature representation, formulated as a local center-of-mass (COM) field, that extracts intrinsic geometric structures, enabling seamless cross-model interaction. Furthermore, we introduce a spikiness-aware selection strategy that evaluates the spatial reliability of each VFM, effectively filtering out high-norm artifacts to aggregate guidance from only the most reliable expert at each local region. DiveUp is a unified, encoder-agnostic framework; a jointly-trained model can universally upsample features from diverse VFMs without requiring per-model retraining. Extensive experiments demonstrate that DiveUp achieves state-of-the-art performance across various downstream dense prediction tasks, validating the efficacy of multi-expert relational guidance. Our code and models are available at the \href{https://github.com/Xiaoqiong-Liu/DiveUp}{DiveUp Repository}.

\end{abstract}

\section{Introduction}
\label{sec:intro}

The pretrained features from vision foundation models (VFMs)~\cite{tschannen2025siglip,simeoni2025dinov3,oquab2023dinov2,ranzinger2024radio} have been widely applied in various vision tasks~\cite{el2024probing,huang2024renovating,kirillov2023segment,fan2025prvql,ravi2024sam,tumanyan2024dino} owing to the rich semantics and excellent generality. Nevertheless, due to patch-based tokenization or pooling operation in these pretrained models, the resulting feature maps are usually much smaller (\eg, 16 times smaller) in spatial resolution than the original input images. This significant reduction in resolution leads to the loss of important spatial details (\eg, geometric structures and object boundaries) in feature maps, thereby limiting their utility for downstream tasks, such as semantic segmentation and depth estimation, that demand pixel-level understanding for dense prediction.

In order to mitigate the issue of low spatial resolution, several recent studies~\cite{couairon2025jafar,fu2024featup,wimmer2026anyup,huang2025loftuplearningcoordinatebasedfeature,chambon2025nafzeroshotfeatureupsampling} have proposed learning-based methods to upsample the features produced by VFMs, and demonstrated promising results. These approaches typically leverage the high-resolution features from the same foundation model to achieve upsampling through self-reconstruction by optimizing a pixel-wise distance (\eg, L2-based or Cosine similarity-based loss). While this paradigm preserves global semantic consistency, relying solely on intra-model features from a single VFM for learning forces the upsampler to overfit to the source model's inherent flaws, such as location misalignment \cite{wang2023sclip}, and high-norm artifacts \cite{darcet2024registers, yang2024denoising}. For instance, the multimodal VFMs (\eg,~\cite{tschannen2025siglip, radford2021learning}) that contrast visual and textual modalities often suffer from severe location misalignment. When the high-resolution features from these models are employed for self-supervision, the reconstructed upsampled features inevitably propagate these similar misalignment issues.

To address the limitations inherent in single-VFM-based upsampling methods, we propose \textbf{DiveUp}, a novel framework that leverages diverse VFMs for feature upsampling. The key motivation behind DiveUp is that different VFMs exhibit distinct representational emphases due to variations in architecture, pretraining objectives, and inductive biases. For example, some models may better capture high-level semantics, while others preserve stronger geometric or structural cues. By treating diverse VFMs as a panel of experts and extracting their structural consensus, the upsampling process can actively regularize against the inherent biases of any single source VFM. However, learning from diverse VFMs is not trivial due to their unaligned feature spaces. To handle this, we formulate a local center-of-mass (COM) field as a universal relational feature representation, which preserves intrinsic geometric structures while enabling cross-model feature interaction. Then, rather than directly using the raw features from VFMs, DiveUp fuses their relational representations to guide the reconstruction of high-resolution features, producing more robust upsampled features. It is worth noting that, our relational feature representation is general and applicable to other existing upsampling methods for improvements. In addition, considering that different VFMs may contribute unequally to different spatial regions, we present a simple yet effective spikiness-aware selection strategy. By dynamically evaluating the spatial reliability of each VFM, this strategy effectively filters out high-norm artifacts and aggregates guidance from only the most reliable expert at each position, further enhancing the feature upsampling quality.

Our DiveUp is a unified, encoder-agnostic framework, and a jointly-trained model can universally upsample features from diverse VFMs without requiring per-model retraining. To validate its effectiveness, we conduct extensive experiments on various downstream dense prediction tasks including semantic segmentation and depth estimation, and the results show that our DiveUp achieves state-of-the-art results and outperforms existing upsamplers, evidencing the efficacy of incorporating diverse VFMs for feature upsampling. Moreover, as an encoder-agnostic and lightweight design, DiveUp offers an efficient plug-and-play solution that enhances VFM backbones without requiring architectural modifications across multiple tasks.

In summary, our contributions are as follows: \textbf{(1)} We introduce DiveUp, a novel framework that for the first time leverages diverse VFMs to provide relational guidance for feature upsampling; \textbf{(2)} We propose a universal relational feature representation, formulated as a local center-of-mass (COM) field, that reconciles unaligned feature spaces to guide the reconstruction of high-resolution features; \textbf{(3)} We present a simple yet effective spikiness-aware selection strategy that filters out high-norm artifacts to dynamically optimize VFM feature fusion; and \textbf{(4)} Extensive experiments show that DiveUp achieves state-of-the-art performance and surpasses existing upsamplers.

\section{Related Work}

Feature upsampling aims to increase the spatial resolution of intermediate feature maps, bridging the gap between coarse patch-level semantics and dense pixel-level prediction tasks. Historically, high-resolution reconstruction is heavily task-specific, relying on end-to-end trained decoders~\cite{Lin2016FeaturePN,Ronneberger2015UNetCN,Chen2018EncoderDecoderWA} tailored for distinct downstream objectives. However, such task-specific decoders inevitably discard the rich, general-purpose semantics of the original VFMs, severely compromising their zero-shot transferability. To preserve these universal representations, recent research has shifted entirely towards task-agnostic feature upsampling, which produces high-resolution, task-generalizable feature spaces. The evolution of this modern field can be broadly categorized into three paradigms.

\paragraph{Classical Handcrafted Upsampling.}
Early non-learning approaches primarily relies on parameter-free interpolation and edge-aware image filtering. Simple approaches like bilinear interpolation operate independently of image content, leading to blurred semantic boundaries. To address this, classical filters such as Joint Bilateral Upsampling (JBU) \cite{kopf2007joint} and Guided Image Filtering (GIF) \cite{He2010GuidedIF} leverage high-resolution RGB images as spatial priors to guide the upsampling of low-resolution signals. While highly efficient and requiring no training, these handcrafted kernels operate fundamentally on low-dimensional spaces (\eg, color or tone) using fixed mathematical forms. Consequently, they struggle to capture the complex, high-dimensional semantic manifolds of modern VFMs.

\paragraph{VFM-Specific Feature Upsampling.}
With the dominance of deep learning, approaches evolve to leverage trainable modules tailored for specific feature spaces. FeatUp~\cite{fu2024featup} and FeatSharp~\cite{Ranzinger2025FeatSharpYV} restore the spatial details by enforcing multi-view consistency. More specifically, FeatUp employs an implicit neural representation, while FeatSharp utilizes a patch-tiling strategy combined with local transformer attention. Opting for a more straightforward and efficient design, LiFT~\cite{suri2024lift} adopts a fully convolutional U-Net architecture for spatial restoration. Taking a different route, LoftUp~\cite{huang2025loftuplearningcoordinatebasedfeature} employs a coordinate-based cross-attention transformer trained in two stages via SAM-generated~\cite{kirillov2023segment} pseudo-labels. Alongside these, the recent JAFAR~\cite{couairon2025jafar} adopts a lightweight cross-attention module that optimizes feature reconstruction at low upsampling ratios to generalize to arbitrary scales. However, the defining bottleneck of these approaches is that they are inherently \textit{VFM-specific}. Their reliance on fixed, model-tied feature spaces for every individual foundation model, fundamentally restricting them from generalizing zero-shot to diverse, unseen VFMs.

\begin{figure}[!t]
    \centering
    \includegraphics[width=0.95\linewidth]{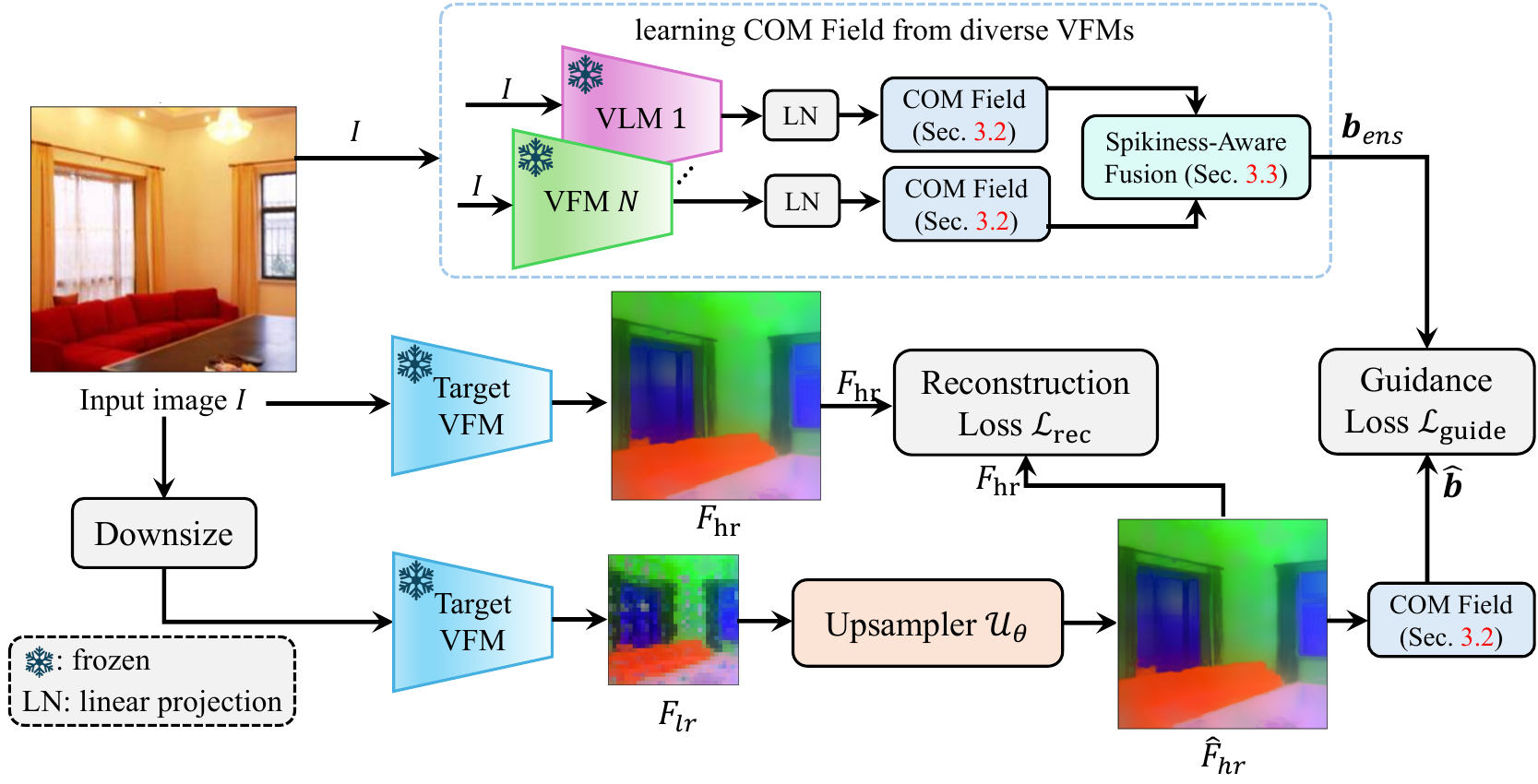}\vspace{-2mm}
    \caption{Overview of the DiveUp, which leverages diverse VFMs to learn feature upsampling. \emph{Best viewed in color and by zooming in for all the figures in this paper}.}
    \label{fig:pipeline}\vspace{-5mm}
\end{figure}

\paragraph{VFM-Agnostic Feature Upsampling.}
To overcome the retraining bottleneck, the community has recently shifted toward universal, feature-agnostic frameworks. Building upon JAFAR's cross-attention baseline, AnyUp~\cite{wimmer2026anyup} introduces a feature-agnostic channel projection layer before the spatial feature transform. This standardizes the input dimension, making the key generation process independent of the specific source VFM. Taking a step further in architectural decoupling, NAF~\cite{chambon2025nafzeroshotfeatureupsampling} completely removes the SFT to isolate the spatial guidance weights from the source features. By discovering that the high-resolution guidance signal is the critical performance bottleneck, NAF relies on a more sophisticated image encoder to drive the cross-scale upsampling. Concurrently, \textit{Upsample Anything}~\cite{seo2025upsample} introduces a test-time adaptation paradigm, bypassing training entirely by optimizing a per-image anisotropic Gaussian kernel that bridges spatial and range cues. When applied to models that suffer from high-norm artifacts or poor dense localization (\eg, text-aligned CLIP), these universal upsamplers merely inherit the input's existing location misalignment. Our DiveUp also belong to the VFM-Agnostic category. However, \textbf{\emph{different from}} the aforementioned methods that rely on a single VFM  for training during the learning process, our DiveUp is able to effectively leverage multiple diverse VFMs by exploiting their complementary representations, leading to significantly more robust upsampled features.

\begin{figure}[t!] 
    \centering
    \includegraphics[width=0.98\textwidth, trim={0.1cm 0.2cm 0.1cm 0.2cm}, clip]{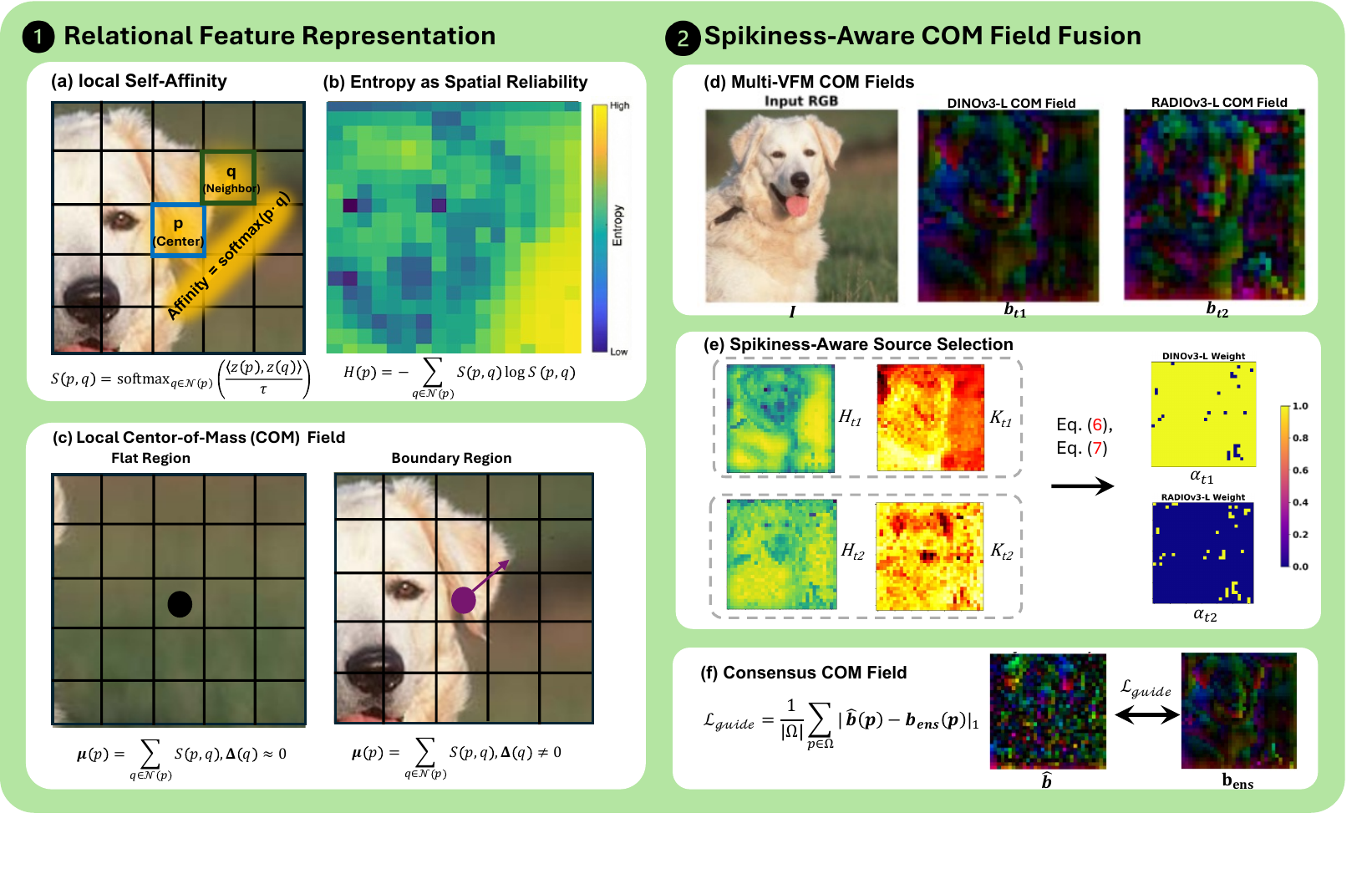} 
    \vspace{-4pt} 
    \caption{The DiveUp Multi-VFM Relational Guidance Framework. 
    \textbf{(1) Relational Feature Representation.} We compute local self-affinity \textbf{(a)} to evaluate spatial reliability via entropy \textbf{(b)} and extract the local center-of-mass (COM) field \textbf{(c)}. In homogeneous regions, the expected spatial offset is negligible ($\boldsymbol{\mu}(p) \approx \mathbf{0}$), whereas it shifts directionally at semantic boundaries ($\boldsymbol{\mu}(p) \neq \mathbf{0}$), effectively capturing intrinsic geometric structures. 
    \textbf{(2) Spikiness-Aware COM Field Fusion.} DiveUp extracts these relational representations from diverse VFMs \textbf{(d)} and employs a \textit{spikiness-aware source selection} \textbf{(e)}. By adaptively evaluating the spatial reliability of each VFM, we generate dynamic gating weights to fuse multiple COM fields into a unified \textit{Consensus COM Field} \textbf{(f)}. This serves as the robust location alignment target for high-resolution feature reconstruction.}
    \label{fig:method_components}
    \vspace{-15pt} 
\end{figure}

\section{The Proposed Methodology}
\label{sec:method}

In this paper, we formulate feature upsampling as a multi-VFM relational guidance task rather than the standard intra-model reconstruction as in existing methods~\cite{couairon2025jafar,wimmer2026anyup,huang2025loftuplearningcoordinatebasedfeature,chambon2025nafzeroshotfeatureupsampling}. Specifically, our DiveUp effectively upsamples low-resolution features by decoupling the optimization into two distinct pathways, including a base semantic reconstruction objective driven by the target VFM, and a geometric structure objective guided by the \emph{Consensus center-of-mass (COM) Field} from an ensemble of diverse VFMs.

\subsection{Problem Formulation and Framework Overview}
\label{subsec:problem_setup}
Given an input image $I \in \mathbb{R}^{H \times W \times 3}$, a vision foundation model (VFM) extracts a low-resolution feature map $F_{lr} \in \mathbb{R}^{h \times w \times C}$. We aim to learn a parameterized neural upsampler $\mathcal{U}_\theta$ to predict a high-resolution feature map $\hat{F}_{hr} \in \mathbb{R}^{H \times W \times d}$ that has the same size of the input, as follows,
\begin{equation}
    \hat{F}_{hr} = \mathcal{U}_\theta(F_{lr}, I). \label{eq:upsampler}
\end{equation}

Inspired by classic edge-preserving filters~\cite{kopf2007joint}, the high-resolution RGB image $I$ is utilized to guide the interpolation. Specifically, following NAF~\cite{chambon2025nafzeroshotfeatureupsampling}, our upsampler $\mathcal{U}_\theta$ is instantiated as a lightweight, spatial-and-content aware interpolation module. Concretely, it employs a compact image encoder to extract high-resolution visual cues from $I$. These cues, augmented with rotary position embeddings (RoPE)~\cite{su2024roformer}, act as spatial queries to dynamically aggregate the low-resolution features $F_{lr}$ via a localized cross-attention mechanism. Due to limited space, we show the architecture of upsampler in the \textbf{supplementary material}.

While existing paradigms optimize the upsampler $\mathcal{U}_\theta$ solely through intra-model self-reconstruction, DiveUp decouples the learning objective into two distinct pathways, including (1) base semantic reconstruction driven by the source VFM, and (2) location alignment driven by a \textit{consensus center-of-mass (COM) field} aggregated from multiple VFMs. Importantly, this \textit{multi-VFM relational guidance} is utilized exclusively during training. During inference, $\mathcal{U}_\theta$ requires no additional heads or computational overhead, serving as an efficient, plug-and-play module for feature upsampling learning as shown in our experiments. Fig.~\ref{fig:pipeline} illustrates the overview of our DiveUp.

\subsection{Relational Feature Representation}
\label{subsec:relational_feature}
Due to unaligned feature spaces of different VFMs, directly fusing their features may degrade the quality. To deal with this issue, we propose to learn a universal relational representation based on local self-affinity for different models.


\paragraph{Local Self-Affinity.}
As illustrated in Fig.~\ref{fig:method_components} (a), for each spatial position $p$, let $\mathcal{N}(p)$ denote a local $w \times w$ window. To reconcile the varying channel dimensions across different VFMs, we first map the raw features $F(p)$ into a common $d$-dimensional space via an independent, VFM-specific projection head $\phi(\cdot)$, yielding projected features $z(p) = \phi(F(p))$. The local similarity distribution is then defined as follows,
\begin{equation}
    S(p,q) = \mathrm{softmax}_{q \in \mathcal{N}(p)}\left( \frac{\langle z(p), z(q) \rangle}{\tau} \right), \quad q \in \mathcal{N}(p).
    \label{eq:affinity}
\end{equation}
where $\langle\cdot,\cdot \rangle$ denotes the dot product. This distribution captures the model's internal structural understanding, indicating which neighboring pixels share semantic affinity with $p$.

\paragraph{Spatial Reliability via Entropy and Spikiness.}
As visualized in Fig.~\ref{fig:method_components}(b), to quantify the reliability of this structural understanding, we compute two complementary metrics. First, the Shannon entropy $H(p)$ of the local affinity measures the spatial uncertainty of boundaries via
\begin{equation}
    H(p) = -\sum_{q \in \mathcal{N}(p)} S(p,q) \log S(p,q). \label{eq:entropy}
\end{equation}
Second, we define feature \textit{spikiness} $K(p)$ to measure the channel-wise activation sharpness, formulated as the ratio between the $L_\infty$ and $L_2$ norms of the local feature vector via
\begin{equation}
    K(p) = \frac{\|z(p)\|_\infty}{\|z(p)\|_2 + \epsilon}. \label{eq:spikiness}
\end{equation}
Critically, extreme channel spikiness typically indicates the presence of global semantic high-norm tokens rather than spatially useful fine-grained details. Therefore, a highly reliable geometric structure is characterized by both low entropy $H(p)$ (confident spatial boundaries) and low spikiness $K(p)$ (absence of artifact interference).

\paragraph{Local Center-of-Mass (COM) Field.}
As shown in Fig.~\ref{fig:method_components}(c), let $\boldsymbol{\Delta}(q) = (\Delta x_q, \Delta y_q)$ be the relative spatial offset of neighbor $q$ in window coordinates. The center-of-mass (COM) vector at position $p$ is computed as the expected spatial offset through
\begin{equation}
    \boldsymbol{\mu}(p) = \sum_{q \in \mathcal{N}(p)} S(p,q)\, \boldsymbol{\Delta}(q). \label{eq:com_vec}
\end{equation}
In geometrically flat regions, $\boldsymbol{\mu}(p) \approx \mathbf{0}$; near semantic boundaries, it heavily biases toward the semantic interior. We normalize this into a bounded vector field $\mathbf{b}(p) = \mathrm{clip}(\boldsymbol{\mu}(p)/r,\,-1,\,1)$ with $r = \lfloor w/2 \rfloor$, forming a universal relational representation that transcends the raw channel dimensions of different VFMs.

\subsection{Spikiness-Aware COM Field Fusion}
\label{subsec:consensus}
As depicted in Fig.~\ref{fig:method_components}(d), for a set of frozen VFMs $\mathcal{T}=\{t_k\}_{i=1}^{N}$ with $t_i$ the $i^{th}$ VFM, we extract features from the same image and resample them to a single COM field. Each model $t_i \in \mathcal{T}$ independently yields its entropy $H_{t_i}$, spikiness $K_{t_i}$, and COM  field $\mathbf{b}_{t_i}$. 

\paragraph{Spikiness-Aware Source Selection.} Considering that different VFMs contribute unequally, rather than simply averaging the unaligned features, we introduce a \textit{spikiness-aware selection strategy} (see Fig.~\ref{fig:method_components}(e)) to dynamically evaluate spatial reliability. To ensure fair comparison across diverse VFMs with varying entropy scales, we first normalize the entropy spatially using standard Z-score normalization for the $i$-th VFM $t_i$, denoted as $\tilde{H}_{t_i}(p)$. 

We then formulate a per-pixel confidence score $g_{t_i}(p)$ that intrinsically favors low spatial uncertainty (negative entropy) while strictly penalizing global high-norm artifacts (excessive spikiness) via
\begin{equation}
    g_{t_i}(p) = -\tilde{H}_{t_i}(p) - \beta \max\bigl(0,\, K_{t_i}(p) - \gamma\bigr), \label{eq:confidence_score}
\end{equation}
where $K_{t_i}(p)$ is the spikiness of the VFM $t_i$ obtained from Eq.~(\ref{eq:spikiness}), $\beta$ controls the penalty strength, and $\gamma$ acts as a tolerance threshold. This hinge-style penalty elegantly suppresses spatially meaningless artifact tokens without disrupting well-behaved geometric features. We apply a winner-take-all hard gating strategy to select the optimal VFM at each spatial location, as follows:
\begin{equation}
    \alpha_{t_i}(p) = \mathbf{1}\bigl[{t_i} = \arg\max_{t' \in \mathcal{T}} g_{t'}(p) \bigr], \label{eq:gating}
\end{equation}
where $\mathbf{1}[\cdot]$ denotes the indicator function, which equals 1 if the condition holds and 0 otherwise.

\paragraph{COM Field Fusion.}
Critically, the gating weights $\alpha_{t_i}(p)$ are \emph{not} used to fuse raw feature maps. Instead, we directly fuse the universal relational representations to form a \textit{Consensus COM Field}. Since $\alpha_{t_i}(p)$ acts as a hard one-hot selector, the fused location alignment target (illustrated in Fig.~\ref{fig:method_components}(f)) is elegantly formulated as follows:
\begin{equation}
    \mathbf{b}_{\mathrm{ens}}(p) = \sum_{t_i \in \mathcal{T}} \alpha_{t_i}(p)\, \mathbf{b}_{t_i}(p). \label{eq:b_ens}
\end{equation}
This consensus field $\mathbf{b}_{\mathrm{ens}}$ dynamically inherits the sharpest and cleanest geometric structures from the most reliable VFM at any given pixel, providing an  robust target to guide the upsampler's location alignment.

\subsection{Learning from Diverse VFMs}
\label{subsec:learning_objectives}
Having fused the diverse geometric structures into a robust consensus field $\mathbf{b}_{\mathrm{ens}}$, we leverage it to supervise the feature upsampling process. Specifically, the total loss $\mathcal{L}_{\mathrm{total}}$ contains two decoupled parts: the intrinsic semantic reconstruction, and the multi-VFM relational guidance using $\mathbf{b}_{\mathrm{ens}}$. Mathematically, $\mathcal{L}_{\mathrm{total}}$ is calculated as follows:
\begin{equation}
    \mathcal{L}_{\mathrm{total}} = \mathcal{L}_{\mathrm{rec}}(\hat{F}_{hr}, F_{hr}) + \lambda\mathcal{L}_{\mathrm{guide}}(\hat{\mathbf{b}}, \mathbf{b}_{\mathrm{ens}}),
    \label{eq:total_loss}
\end{equation}
where the base reconstruction loss $\mathcal{L}_{\mathrm{rec}}$ computes the mean squared error (MSE) between the predicted high-resolution feature $\hat{F}_{hr}$ and the target high-resolution crop feature $F_{hr}$ to maintain basic semantic identity, and $\lambda$ serves as the balancing weight. The relational guidance loss $\mathcal{L}_{\mathrm{guide}}$ is specifically designed to enforce the location alignment provided by the VFM consensus, and is computed as follows:
\begin{equation}
    \mathcal{L}_{\mathrm{guide}} = \frac{1}{|\Omega|} \sum_{p \in \Omega} \bigl\| \hat{\mathbf{b}}(p) - \mathbf{b}_{\mathrm{ens}}(p) \bigr\|_1.
    \label{eq:guide_loss}
\end{equation}
Specifically, $\mathcal{L}_{\mathrm{guide}}$ aligns the upsampler's predicted COM field $\hat{\mathbf{b}}$ to the fused VFM-consensus target $\mathbf{b}_{\mathrm{ens}}$ via an $L_1$ distance across the spatial domain, where $\Omega$ denotes the set of all spatial locations in the high-resolution feature map and $|\Omega|$ is the total number of spatial positions.

\section{Experiments}
\label{sec:exp}

We assess the effectiveness of DiveUp across multiple VFMs. Following previous methods~\cite{couairon2025jafar,wimmer2026anyup,chambon2025nafzeroshotfeatureupsampling}, we conduct linear probing on the upsampled features and assess their performance on different dense prediction tasks, including semantic segmentation and depth estimation.

\subsection{Experimental Setup}
\label{sec:exp_setup}

\paragraph{Implementation Details.} DiveUp is implemented using PyTorch~\cite{paszke2019pytorch} on a single NVIDIA RTX A6000 GPU. Following~\cite{chambon2025nafzeroshotfeatureupsampling}, we train the upsampler in DiveUp for 25K iterations on the ImageNet~\cite{deng2009imagenet} dataset, using the AdamW optimizer~\cite{loshchilov2017decoupled} with a batch size of 2, an initial learning rate of 2e-4, and a weight decay of 1e-5. The local window size for computing the self-affinity and COM field is set to $w=7$. To learn a robust VFM-agnostic upsampler, we employ a mixed-training strategy that simultaneously applies four backbones during training: SigLIP-B~\cite{tschannen2025siglip}, DINOv3-B~\cite{simeoni2025dinov3}, DINOv2-S~\cite{oquab2023dinov2}, and RADIOv2.5~\cite{ranzinger2024radio}. For each backbone, the upsampler is guided by relational representations extracted from an ensemble of two diverse VFMs: DINOv3-L~\cite{simeoni2025dinov3} and RADIOv3-L~\cite{heinrich2025radiov25}. In the training objective, the guidance weight $\lambda$ is empirically set to 0.5. For the spikiness-aware selection strategy, the tolerance threshold $\gamma$ is set to 0.6, and the penalty weight $\beta$ is set to 20.0. Notably, DiveUp is an elegant single-stage framework with a training time of less than 2 hours, entirely avoiding the computationally expensive high-resolution refinement stage required by prior works. Our code and models will be released.

\paragraph{Linear Probing Protocol and Tasks.} 
To assess DiveUp, following~\cite{couairon2025jafar,chambon2025nafzeroshotfeatureupsampling}, we conduct linear probing of upsampled features and evaluate their performance on two representative dense prediction tasks, including semantic segmentation and depth estimation. Similar to~\cite{chambon2025nafzeroshotfeatureupsampling}, the VFM receives the normalized 448$\times$448 image as input, and the extracted feature is upsampled to the original image resolution, corresponding to $\times 14$~\cite{oquab2023dinov2} or $\times 16$~\cite{simeoni2025dinov3, tschannen2025siglip} spatial scaling. Then, a linear classifier with a 1$\times$1 convolution layer is trained on top of these upsampled features to predict per-pixel semantic labels or depth values. For VFM-specific methods~\cite{couairon2025jafar,huang2025loftuplearningcoordinatebasedfeature}, we use their official implementations on each corresponding VFM as in~\cite{chambon2025nafzeroshotfeatureupsampling} and then perform linear probing.

\renewcommand{\arraystretch}{1.25}
\begin{table}[!t]
\centering
\setlength{\tabcolsep}{2pt}
\caption{Linear probing results of semantic segmentation (mIoU $\uparrow$) on Pascal VOC~\cite{everingham2010pascal} and depth estimation ($\delta_1 \uparrow$) on NYUv2~\cite{silberman2012indoor}. We evaluate features extracted from various VFMs. `V.A.' indicates VFM-agnostic models. `${\dagger}$' denotes results of NAF~\cite{chambon2025nafzeroshotfeatureupsampling} are obtained using an additional refinement stage. `-' indicates results not available. The best and the second best results are highlighted in \textbf{bold} and \underline{underlined}, respectively, for all tables in the paper.}\vspace{-2mm}
\label{tab:main_results_combined}
\resizebox{0.99\textwidth}{!}{%
\begin{tabular}{rccccccccc}
\Xhline{1.2pt}
\multirow{2}{*}{Methods} & \multirow{2}{*}{V.A.} & \multicolumn{4}{c}{Semantic Segmentation (Pascal VOC)} & \multicolumn{4}{c}{Depth Estimation (NYUv2)} \\ 
\cmidrule(lr){3-6} \cmidrule(lr){7-10}
 & & SigLIP-B & DINOv2-S & RADIOv2.5 & DINOv3-B & SigLIP-B & DINOv2-S & RADIOv2.5 & DINOv3-B \\ 
 \hline\hline
Bilinear & \ding{51} & 71.88 & 80.70 & 84.46 & 86.99 & 71.49 & 82.51 & 83.06 & 85.39 \\
LoftUp\textcolor{gray}{\scriptsize{[ICCV'25]}}~\cite{huang2025loftuplearningcoordinatebasedfeature}& \ding{55} & 46.77 & 83.66 & - & - & 48.54 & 80.85 & - & -\\
JAFAR\textcolor{gray}{\scriptsize{[NeurIPS'25]}}~\cite{couairon2025jafar} & \ding{55} & 77.36 & 84.24 & 85.93 & 87.10 & 75.50 & 83.40 & 83.51 & 84.38 \\
AnyUp-S\textcolor{gray}{\scriptsize{[ICLR'26]}}~\cite{wimmer2026anyup} & \ding{51} & \underline{78.48} & 83.85 & 85.51 & 86.62 & 74.46 & 83.61 & 84.83 & 86.30 \\
AnyUp-M\textcolor{gray}{\scriptsize{[ICLR'26]}}~\cite{wimmer2026anyup} & \ding{51} & 77.71 & 84.11 & 85.67 & 86.14 & \underline{75.81} & 83.70 & \underline{84.93} & \underline{86.54} \\
NAF\textcolor{gray}{\scriptsize{[CVPR'26]}}~\cite{chambon2025nafzeroshotfeatureupsampling}$^{\dagger}$ & \ding{51} & 77.79 & \underline{84.52} & \underline{86.60} & \textbf{87.85} & 75.20 & \underline{84.06} & 83.88 & 85.63 \\ \hline
\rowcolor[HTML]{E7E7FF}
DiveUp (ours) & \ding{51} & \textbf{80.78} & \textbf{85.08} & \textbf{87.07} & \underline{87.62} & \textbf{77.77} & \textbf{84.73} & \textbf{85.50} & \textbf{86.79} \\
\Xhline{1.2pt}
\\
\end{tabular}%
}
\end{table}

\renewcommand{\arraystretch}{1.0}
\begin{table}[!t]
\centering
\setlength{\tabcolsep}{9pt}
\caption{Comparison across tasks on different datasets. The linear probing results of all methods are obtained using DINOv2-S. `V.A.' indicates VFM-agnostic models. `${\dagger}$' denotes results of NAF~\cite{chambon2025nafzeroshotfeatureupsampling} are obtained using an additional refinement stage.}\vspace{-2mm}
\label{tab:across_datasets}
\resizebox{0.85\textwidth}{!}{%
\begin{tabular}{rccccc}
\Xhline{1.2pt}
 & & \multicolumn{3}{c}{Semantic Segmentation (mIoU $\uparrow$)} & Depth Est. ($\delta_1 \uparrow$) \\ \cmidrule(lr){3-5} \cmidrule(lr){6-6}
\multirow{-2}{*}{Methods} & \multirow{-2}{*}{V.A.} & COCO & Pascal VOC & ADE20K & NYUv2  \\ \hline\hline
Bilinear & \ding{51} & 59.03 & 80.70 & 39.23 & 82.51 \\
FeatUp\textcolor{gray}{\scriptsize{[ICLR'24]}}~ \cite{fu2024featup} & \ding{55} & 60.10 & 81.08 & 38.82 & 82.53 \\
LoftUp\textcolor{gray}{\scriptsize{[ICCV'25]}}~\cite{huang2025loftuplearningcoordinatebasedfeature} & \ding{55} & \underline{62.23} & 83.66 & 42.01 & 80.85 \\
JAFAR\textcolor{gray}{\scriptsize{[NeurIPS'25]}}~\cite{couairon2025jafar} & \ding{55} & 60.78 & 84.24 & 40.49 & 83.40 \\
AnyUp-S\textcolor{gray}{\scriptsize{[ICLR'26]}}~\cite{wimmer2026anyup} & \ding{51} & 61.10 & 83.85 & 42.26 & 83.61 \\
NAF\textcolor{gray}{\scriptsize{[CVPR'26]}}~\cite{chambon2025nafzeroshotfeatureupsampling}$^{\dagger}$ & \ding{51} & 62.18 & \underline{84.52} & \underline{42.70} & \underline{84.06} \\ 
\Xhline{1.2pt}
\rowcolor[HTML]{E7E7FF}
DiveUp (ours) & \ding{51} & \textbf{62.71} & \textbf{85.08} & \textbf{42.82} & \textbf{84.73} \\ \hline
\end{tabular}%
}
\end{table}

\begin{figure}[!t]
    \centering
    \includegraphics[width=0.95\linewidth]{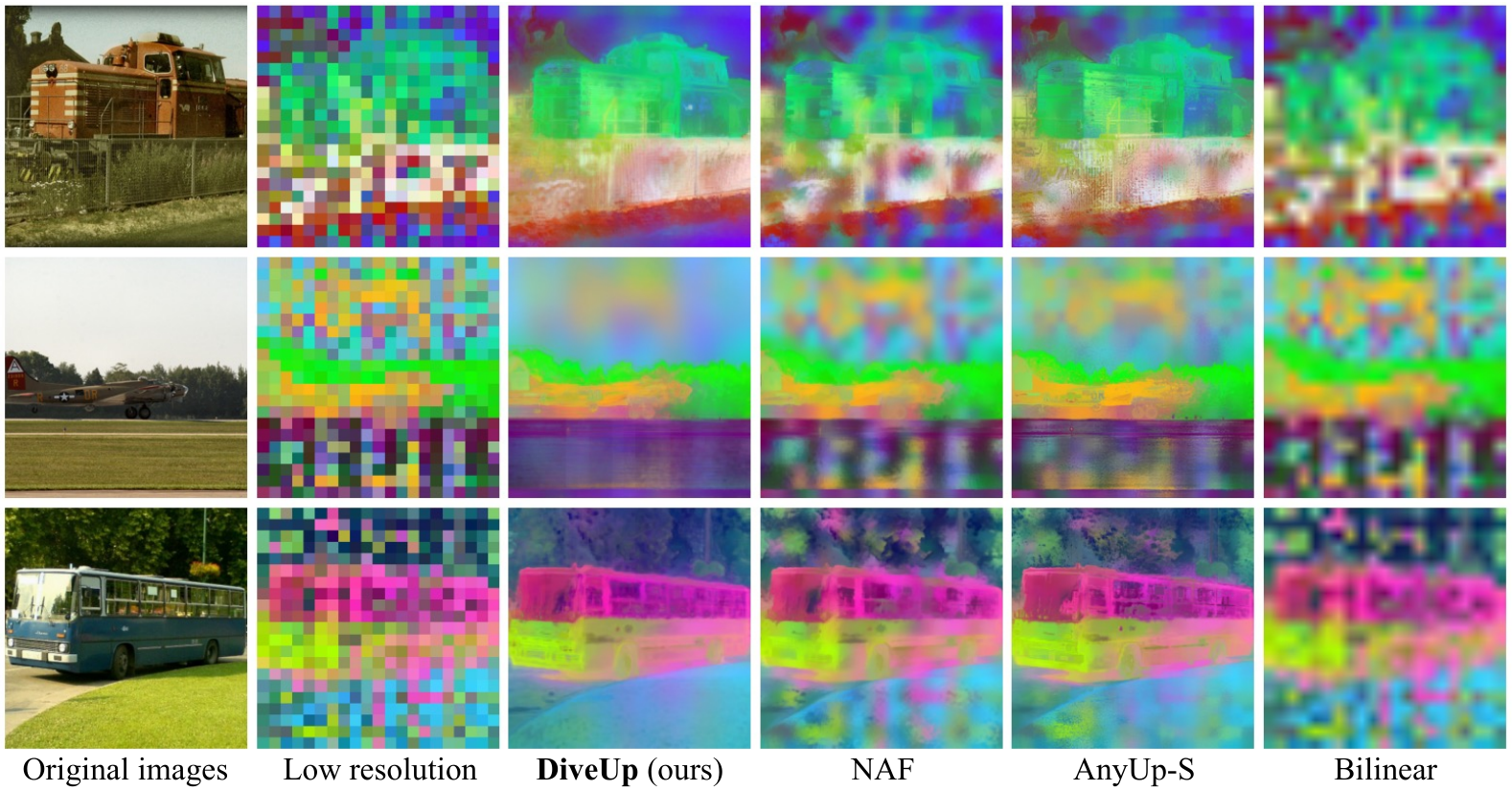}\vspace{-2mm}
    \caption{Comparison of upsampled features from different methods.}
    \label{fig:feat_qua}\vspace{-5mm}
\end{figure}

\subsection{State-of-the-art Comparison}

\paragraph{Comparison Across VFMs.} We compare DiveUp against both VFM-specific upsamplers~\cite{couairon2025jafar, fu2024featup, huang2025loftuplearningcoordinatebasedfeature} and VFM-agnostic methods~\cite{wimmer2026anyup,chambon2025nafzeroshotfeatureupsampling}. To evaluate the generality of different methods, we evaluate them across a diverse set of representative vision foundation models, including SigLIP-B, DINOv2-S, RADIOv2.5, and DINOv3-B. Experiments are conducted on semantic segmentation using Pascal VOC~\cite{everingham2010pascal} and on depth estimation using NYUv2~\cite{silberman2012indoor}.

Tab.~\ref{tab:main_results_combined} reports the comparison results. It is worth noting that, the method of NAF achieves feature unsampling through a two-stage framework, in which the second stage is to perform refinement on high-resolution 1,024$\times$1,024 with DINOv3-B. Despite this, from  Tab.~\ref{tab:main_results_combined}, we can observe that, our DiveUp achieves the best results on 3 out of 4 VFMs for semantic segmentation with 80.78\% mIoU on SigLIP-B, 85.08\% mIoU on DIVOv2-S, 87.07\% mIoU on RADIOv2.5, clearly outperforming other methods. On the large-scale DINOv3-B, NAF achieves the best result with 87.85\% mIoU via two stages. Despite lacking the refinement stage, our DiveUp achieves comparable performance with 87.62\% mIoU, showing its effectiveness. For depth estimation, DiveUp achieves the best results on all VFMs with 77.77\% $\delta_1$ score on SigLIP-B, 84.73\% $\delta_1$ score on DINOv2-S, 85.50\% $\delta_1$ score on RADIOv2.5, and 86.79\% $\delta_1$ score on DINOv3-B, clearly surpassing all other methods.

\paragraph{Comparison Across Tasks and Datasets.} We further assess and compare different methods on different tasks and datasets using DINOv2-S~\cite{oquab2023dinov2}, and Tab.~\ref{tab:across_datasets} shows the results. For semantic segmentation on three benchmarks, including COCO~\cite{lin2014microsoft}, Pascal VOC~\cite{everingham2010pascal}, and ADE20K~\cite{zhou2017scene}, DiveUp consistently achieves the best performance with 62.71\%, 85.08\%, and 42.82\% mIoU scores on COCO, Pascal VOC, and ADE20K, outperforming all other methods on three datasets. In addition, for depth estimation, DiveUp obtains the best result with 84.73 $\delta_1$ score. All these comparison results show the robustness and efficacy of our method.

\paragraph{Qualitative Comparison on Upsampled Features.} To qualitatively compare different methods, we visualize the upsampled features from the SigLIP-B using principal component analysis (PCA)~\cite{mackiewicz1993principal}. We choose SigLIP-B for upsampled feature comparison because SigLIP-B is a visual-textual multimodal model and often yields spatially ambiguous raw features. As shown in Fig.~\ref{fig:feat_qua}, existing upsamplers relying on a single VFM suffer from severe semantic leakage (\eg, the halo effect around the airplane or the red feature of the bus bleeding into the surrounding grass). Owing to the complementary information from diverse VFMs, our DiveUp can alleviate this issue and generate high-quality upsampled features. 

\begin{figure}[!t]
    \centering
    \includegraphics[width=0.95\linewidth]{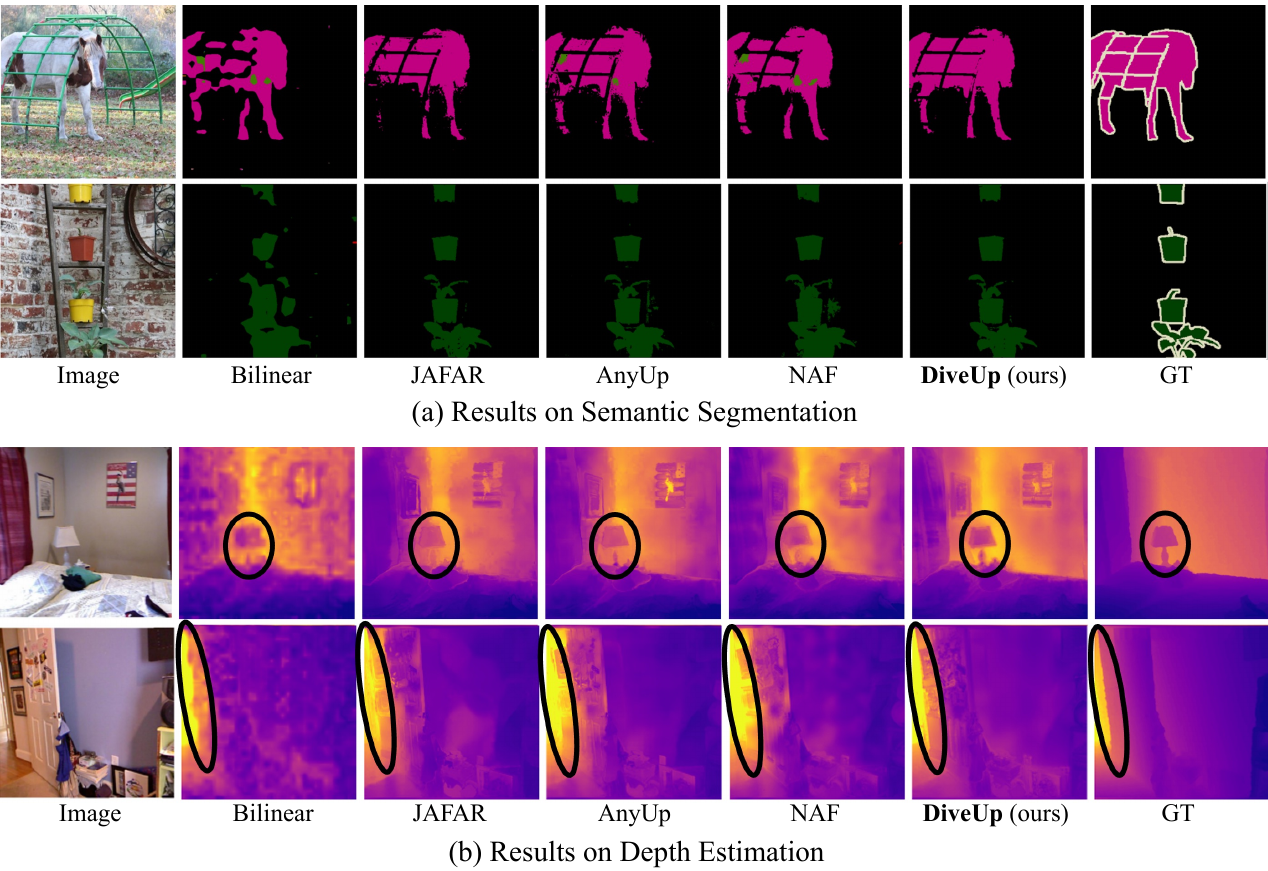}\vspace{-2mm}
    \caption{Comparison of prediction results on semantic segmentation (a) in Pascal VOC and depth estimation (b) for different methods in NYUv2.}
    \label{fig:pred_qua}\vspace{-5mm}
\end{figure}

\paragraph{Qualitative Comparison on Prediction Results.} To further demonstrate and compare different methods, we show the prediction results of upsamplers on semantic segmentation and depth estimation in Fig.~\ref{fig:pred_qua}. From Fig.~\ref{fig:pred_qua}, we can observe that, our DiveUp is able to produce better prediction results than other existing upsamplers, validating the importance and efficacy of learning from diverse VFMs.

\subsection{Zero-Shot Generalization on Unseen Semantic Spaces}

\setlength{\columnsep}{10pt}
\setlength\intextsep{0pt}
\begin{wraptable}{r}{0.5\textwidth} 
\centering
\setlength{\tabcolsep}{5pt}
\caption{Generalization to unseen semantic spaces on Cityscapes~\cite{cordts2016cityscapes} using mIoU ($\uparrow$).}
\label{tab:generalization}
\resizebox{\linewidth}{!}{%
\begin{tabular}{rccc}
\Xhline{1.2pt}
 & \multicolumn{2}{c}{Target VFMs} &  \\ \cline{2-3}
\multirow{-2}{*}{Methods}  & CLIP-B & SwinT-L & \multirow{-2}{*}{Avg.} \\ \hline\hline
NAF\textcolor{gray}{\scriptsize{[CVPR'26]}}~\cite{chambon2025nafzeroshotfeatureupsampling}  & \underline{53.27} & \underline{50.83} & \underline{52.05} \\
AnyUp-S\textcolor{gray}{\scriptsize{[ICLR'26]}}~\cite{wimmer2026anyup}& 50.89 & 49.90 & 50.40 \\ 
AnyUp-M\textcolor{gray}{\scriptsize{[ICLR'26]}}~\cite{wimmer2026anyup} & 50.43 & 48.96 & 49.70 \\ \hline
\rowcolor[HTML]{E7E7FF}
DiveUp (ours) & \textbf{53.65} & \textbf{52.58} & \textbf{53.12} \\ \Xhline{1.2pt}
\end{tabular}
}\vspace{5pt}
\end{wraptable}
To evaluate the zero-shot generalization, we conduct experiments by applying upsamplers on target foundation models representing entirely different semantic and structural spaces, including text-aligned CLIP~\cite{radford2021learning} and window-attention SwinT~\cite{liu2021swin}. As shown in Table~\ref{tab:generalization}, single-backbone methods, NAF and AnyUp-S, suffer degradation on unseen feature spaces. Surprisingly, AnyUp-M exhibits severe \textit{negative transfer}; despite explicitly observing CLIP during training, its performance drops below its single-backbone counterpart. This reveals a critical flaw in existing paradigms: naively mixing diverse VFMs under a self-reconstruction objective dilutes the upsampler's capacity and causes feature confusion. In contrast, DiveUp consistently establishes the new state-of-the-art. Notably, it outperforms AnyUp-M on CLIP-B in a strictly \textit{zero-shot} setting, completely overriding the latter's ``in-domain'' advantage. Furthermore, DiveUp achieves a massive +1.75\% mIoU margin over NAF on Swin-L, proving its superior ability to correct complex artifacts unique to specific models. This rigorously confirms that our relational representation is feature-agnostic rather than merely overfitting to specific models.

\subsection{Generality and Plug-and-Play Capability}

\setlength{\columnsep}{10pt}
\begin{wraptable}{r}{0.45\textwidth}
\centering
\setlength{\tabcolsep}{10pt}
\caption{Generality of DiveUp in existing upsamplers JAFAR and NAF. The experiments are conducted on semantic segmentation with DINOv2-S on Pascal VOC. * indicates re-trained results for fair comparison.}
\label{tab:plugin}
\resizebox{\linewidth}{!}{%
\begin{tabular}{rc}
\Xhline{1.2pt}
Methods & Pascal VoC (mIoU $\uparrow$)   \\ \hline\hline
JAFAR\textcolor{gray}{\scriptsize{[NeurIPS'25]}}~\cite{couairon2025jafar}* & 84.15 \\
\rowcolor[HTML]{E7E7FF} 
JAFAR with DiveUp  & 84.94 {\small \textcolor{blue}{(+0.79)}}  \\ \hline
NAF\textcolor{gray}{\scriptsize{[CVPR'26]}}~\cite{chambon2025nafzeroshotfeatureupsampling}*  & 84.29  \\
\rowcolor[HTML]{E7E7FF} 
NAF with DiveUp  & 84.92 {\small \textcolor{blue}{(+0.63)}}  \\ \Xhline{1.2pt}
\end{tabular}%
}
\vspace{5pt}
\end{wraptable}

DiveUp aims to improve feature upsampling with multi-VFM relational guidance. To validate the generality of this mechanism, we evaluate it as a plug-and-play supervision signal within existing state-of-the-art frameworks, specifically JAFAR and NAF. In this setup, all models are trained and probed using the identical DINOv2-S backbone to ensure a strict and fair comparison. The baseline models are optimized solely with their default self-reconstruction objectives, whereas the enhanced versions (denoted as "with DiveUp") incorporate our relational guidance during training. Tab.~\ref{tab:plugin} reports the results on Pascal VOC. As shown in Table~\ref{tab:plugin}, injecting this structural prior yields consistent improvements. Specifically, the mIoU scores of JAFAR and NAF are significantly boosted by 0.79\% and 0.63\%, respectively, demonstrate that our multi-VFM consensus is not restricted to a specific architecture, but serves as a universally applicable training paradigm. 

\subsection{Efficacy and Complexity Analysis}

To comprehensively evaluate our method under a challenging $\times16$ upsampling ratio ($28^2 \rightarrow 448^2$), we compare its computational metrics (estimated on a single RTX A6000 GPU with a batch size of 2) and downstream DINOv2-S probing performance against state-of-the-art methods. As shown in Tab.~\ref{tab:efficiency}, DiveUp shares an identical lightweight inference profile with NAF due to their similar upsampler architectures. Crucially, it significantly outperforms all other competitors, including computationally expensive models like JAFAR and AnyUp. Furthermore, DiveUp features an elegant one-stage training framework; although trained with multiple VFMs simultaneously, it remains vastly more training efficient than methods like FeatUp (23.2h) and AnyUp (10.1h).

\begin{table}[!t]
\setlength{\tabcolsep}{8pt}
\centering
\caption{Comparison on model efficacy and complexity for different methods for a $\times 16$ upsampling of input features}\vspace{-2mm}
\label{tab:efficiency}
\resizebox{0.95\linewidth}{!}{%
\begin{tabular}{rccccccc}
\Xhline{1.2pt}
Methods & \tabincell{c}{Params \\(M $\downarrow$)} &\tabincell{c}{Tra. Time \\(hour $\downarrow$)} & \tabincell{c}{FLOPs \\ (G $\downarrow$)} & \tabincell{c}{Inf. Speed \\ (FPS $\uparrow$)} & \tabincell{c}{Inf. GPU  \\ Mem. (G $\downarrow$)} & \tabincell{c}{Pascal VOC \\ (mIoU $\uparrow$)} & \tabincell{c}{NYUv2 \\($\delta_1$ $\uparrow$)} \\ \hline\hline
FeatUp\textcolor{gray}{\scriptsize{[ICLR'24]}}~\cite{fu2024featup} & 0.17 & 23.2 & 83.0 & 17.5 & 2.6 & 81.08 & 82.53 \\ 
JAFAR\textcolor{gray}{\scriptsize{[NeurIPS'25]}}~\cite{couairon2025jafar} & 0.63 & 2.7 & 366.5 & 9.2 & 5.3 & 84.24 & 83.40\\
AnyUp\textcolor{gray}{\scriptsize{[ICLR'26]}}~\cite{wimmer2026anyup} & 0.88 & 10.1 & 328.5 & 5.5 & 6.0 & 83.85 & 83.61 \\ 
NAF\textcolor{gray}{\scriptsize{[CVPR'26]}}~\cite{chambon2025nafzeroshotfeatureupsampling} & 0.66 & 1.0 & 265.1 & 17.8 & 1.7 & 84.52 & 84.06 \\
\hline
\rowcolor[HTML]{E7E7FF} 
\textbf{DiveUp (ours)} & 0.66 & 1.5 & 265.1 & 17.8 & 1.7 & 85.08 & 84.73 \\ \Xhline{1.2pt}
\end{tabular}\vspace{-5mm}
}
\end{table}

\subsection{Ablation Studies}
\label{sec:ablations}

To better understand our DiveUp, we conduct ablation studies on semantic segmentation using Pascal VOC. Our final configuration is highlighted in {\color{cyan!75} cyan}

\setlength{\columnsep}{10pt}
\begin{wraptable}{r}{0.5\textwidth}
\centering
\setlength{\tabcolsep}{7pt}
\caption{Ablation experiments on applying different VFMs in DiveUp using frozen CLIP as $F_{lr}$. }
\label{tab:abla_vfm}
\resizebox{0.97\linewidth}{!}{%
\begin{tabular}{rccc}
\Xhline{1.2pt}
          & DINOv3-L & RADIOv3-L & Pascal VOC (mIoU $\uparrow$) \\
    \hline\hline
    \ding{182}     &       &       & 76.42 \\
    \ding{183}     & \ding{51}     &       & 77.18 \textcolor{blue}{(+0.76)} \\
    \ding{184}     &       & \ding{51}     & 77.57 \textcolor{blue}{(+1.15)}\\
    \rowcolor{cyan!10} 
    \ding{185}     & \ding{51}     & \ding{51}     & 77.84 \textcolor{blue}{(+1.42)} \\
    \Xhline{1.2pt}
    \end{tabular}%
}
\vspace{2pt}
\end{wraptable}
\paragraph{Ablation on VFMs.} Our DiveUp utilizes two excellent VFMs, including DINOv3-L and RADIOv3-L, for feature upsampling. To study their impact, we conduct an ablation experiment and the results are shown in Tab.~\ref{tab:abla_vfm}. From Tab.~\ref{tab:abla_vfm}, we can observe that, compared with the original performance, when guided by DINOv3-L or RADIOv3-L alone in DiveUp, yields a 0.76\% and  1.15\% gains, correspondingly. (\ding{182} \emph{v.s.} \ding{184}). These results indicate that, using an additional VFM, even alone, the upsampled features can be improved. When applying both DINOv3-L and RADIOv3-L together, we achieve the best performance with 1.42\% gains (\ding{182} \emph{v.s.} \ding{185}), evidencing the efficacy of diverse VFMs for improving feature upsampling.

\setlength{\columnsep}{10pt}
\begin{wraptable}{r}{0.5\textwidth}
\centering
\setlength{\tabcolsep}{7pt}
\caption{Ablation experiments on VFM fusion strategies in DiveUp  using frozen DINOv2-S as $F_{lr}$.}
\label{tab:abla_sa}
\resizebox{0.97\linewidth}{!}{%
\begin{tabular}{ccc}
\Xhline{1.2pt}
          & VFM Fusion & Pascal VOC (mIoU $\uparrow$) \\ \hline\hline
    \ding{182}     & Addition &  84.51 \\
    \ding{183}     & Concatenation & 84.66 \\
    \rowcolor{cyan!10} 
    \ding{184}     & SA selection (ours) & 84.92 \\
    \Xhline{1.2pt}
    \end{tabular}%
}
\vspace{2pt}
\end{wraptable}
\paragraph{Ablation on VFM Fusion Method.} Considering that, 
different VFMs may contribute unequally to different spatial regions, we present
a simple yet effective spikiness-aware (SA) selection strategy for VFM feature fusion
by aggregating features from the most reliable VFM at each position. To validate its effectiveness, we conduct an ablation study in Tab.~\ref{tab:abla_sa} by comparing it with the simple addition or concatenation methods for VFM fusion. As shown in Tab.~\ref{tab:abla_sa}, we can observe that, DiveUp with our SA selection achieves the best performance and outperforms other fusion strategies, showing its efficacy for improving feature upsampling.

Due to space limitations, more extensive visualizations, ablation studies, and detailed experimental analyses are provided in the \textbf{supplementary material}.

\section{Discussion and Conclusion}
\label{sec:conclusion}

\textbf{Discussion.} Our relational guidance leverages the consensus COM field, which provides superior location alignment compared to the native feature spaces of many target VFMs. This mechanism is particularly beneficial for multimodal or window-attention-based architectures, such as SigLIP~\cite{tschannen2025siglip} and Swin Transformer~\cite{liu2021swin}, which often suffer from severe location misalignment or structural artifacts. DiveUp effectively corrects these misalignments via the proposed \textit{Guidance Loss}, ensuring accurate spatial correspondence during the upsampling process. Conversely, for scale-invariant architectures like DINOv3~\cite{simeoni2025dinov3} that inherently possess well-aligned, dense spatial structures even at low resolutions, the geometric gain is naturally less pronounced. For such highly aligned backbones, current learning-based upsampling methods, including DiveUp, perform comparably to or only marginally better than simple bilinear interpolation. Furthermore, because DiveUp relies on clean RGB images to formulate spatial queries, it may be vulnerable to severe image-level noise. 

\vspace{0.3em}
\noindent
\textbf{Conclusion.} In this paper, we introduce DiveUp, which leverages diverse VFMs for feature upsampling by exploiting their complementary representations. To handle unaligned feature spaces across different VFMs, we propose a simple yet effective relational feature representation that preserves intrinsic geometric structures while enabling cross-model feature interaction. Instead of directly using raw features from VFMs, DiveUp fuses their relational representations to guide the reconstruction of high-resolution features, achieving more robust feature upsampling. In addition, we introduce a simple yet effective spikiness-aware selection strategy for VFM feature fusion by aggregating features from the most reliable VFM at each position, further enhancing the performance. DiveUp is a unified framework, which can universally upsample features from diverse VFMs without per-model retraining. Extensive experiments validate the effectiveness of DiveUp.

%
%
\bibliographystyle{splncs04}
\bibliography{main}
\clearpage
\appendix

\section{Extended Quantitative Evaluations}

\subsection{Zero-Shot Generalization on Noisy Backbones}
In the main paper, we evaluate DiveUp's cross-VFM generalization on standard vision foundation models (e.g., CLIP~\cite{radford2021learning}, Swin~\cite{liu2021swin}). To further evaluate its robustness, we test DiveUp in a strictly zero-shot setting against highly diverse and noisy feature spaces (e.g., multimodal features from PaliGemma2~\cite{Steiner2024PaliGemma2A}, or text-aligned spaces in DFN-CLIP~\cite{fang2023data}). 

As shown in Table~\ref{tab:supp_zeroshot}, while existing single-backbone reconstruction methods (NAF~\cite{chambon2025nafzeroshotfeatureupsampling}, AnyUp-S~\cite{wimmer2026anyup}) suffer degradation when confronted with unseen noisy feature distributions, DiveUp consistently establishes the state-of-the-art across all evaluated architectures. This empirically validates that our multi-VFM relational guidance captures intrinsic geometric structures, making it resilient to various feature noise.

\renewcommand{\arraystretch}{1.25}
\vspace{10pt}
\begin{table}[htbp]
\centering
\caption{Zero-Shot Generalization Across Diverse and Noisy VFMs. Performance evaluated on the PASCAL VOC segmentation benchmark using mIoU (\%) without per-model retraining. DiveUp shows robustness and architecture-agnostic generalization, outperforming baselines across all evaluated noisy backbones. The best and the second best results are highlighted in \textbf{bold} and \underline{underlined}, respectively.}\vspace{-2mm}
\label{tab:supp_zeroshot}
\resizebox{\textwidth}{!}{%
\begin{tabular}{lcccc|cc}
\Xhline{1.2pt}
\multirow{2}{*}{\textbf{Target VFM}} & \multicolumn{4}{c|}{\textbf{Semantic Segmentation (mIoU \%)}} & \multicolumn{2}{c}{\textbf{Absolute Gains}} \\ \cmidrule(lr){2-5} \cmidrule(lr){6-7}
 & NAF~\cite{chambon2025nafzeroshotfeatureupsampling} & AnyUp-S~\cite{wimmer2026anyup} & AnyUp-M~\cite{wimmer2026anyup} & \cellcolor[HTML]{E7E7FF}\textbf{DiveUp (Ours)} & \textbf{vs. NAF} & \textbf{vs. AnyUp-M} \\ 
\hline\hline
ResNet-50 \cite{he2016deep}     & 51.44 & 54.60 & \underline{56.85} & \cellcolor[HTML]{E7E7FF}\textbf{57.08} & \textcolor{blue}{+5.64} & \textcolor{blue}{+0.23} \\
PE-Core-S \cite{bolya2025perception}     & 72.35 & 73.69 & \underline{73.96} & \cellcolor[HTML]{E7E7FF}\textbf{75.31} & \textcolor{blue}{+2.96} & \textcolor{blue}{+1.35} \\
DFN-CLIP-B \cite{fang2023data} & 77.66 & 77.62 & \underline{78.11} & \cellcolor[HTML]{E7E7FF}\textbf{78.81} & \textcolor{blue}{+1.15} & \textcolor{blue}{+0.70} \\
PaliGemma2  \cite{Steiner2024PaliGemma2A}    & 79.58 & 79.64 & \underline{80.24} & \cellcolor[HTML]{E7E7FF}\textbf{81.33} & \textcolor{blue}{+1.75} & \textcolor{blue}{+1.09} \\
MaskCLIP-B \cite{zhou2022maskclip}     & 73.37 & 73.75 & \underline{75.14} & \cellcolor[HTML]{E7E7FF}\textbf{75.57} & \textcolor{blue}{+2.20} & \textcolor{blue}{+0.43} \\
ClearCLIP-B \cite{lan2024clearclip}   & 73.79 & 75.37 & \underline{75.95} & \cellcolor[HTML]{E7E7FF}\textbf{76.19} & \textcolor{blue}{+2.40} & \textcolor{blue}{+0.24} \\
\Xhline{1.2pt}
\end{tabular}%
}
\end{table}

\subsection{Plug-and-Play Robustness with Noisy VFMs}
In the main paper, we demonstrate that DiveUp can be integrated into existing architectures (e.g., JAFAR~\cite{couairon2025jafar} and NAF~\cite{chambon2025nafzeroshotfeatureupsampling}) to enhance upsampling quality on standard VFMs. Here, we evaluate this plug-and-play capability on noisy and unaligned backbones. 

We adopt NAF~\cite{chambon2025nafzeroshotfeatureupsampling} as the baseline architecture, which relies on self-reconstruction. We replace its native intra-model objective with our multi-VFM relational guidance (denoted as NAF w/ DiveUp). As shown in Table~\ref{tab:supp_plug_play}, the baseline NAF struggles to extract coherent spatial structures from these noisy features. However, injecting our relational guidance yields consistent improvements across all evaluated models. Notably, on CLEAR-CLIP and PE-Core-S, our guidance provides absolute gains of \textbf{+2.72\%} and \textbf{+1.54\%} mIoU, respectively. This establishes that our \textit{Consensus COM Field} is a robust geometric prior that can effectively improve baseline upsampler architectures in noisy feature spaces.

\renewcommand{\arraystretch}{1.25}
\begin{table}[htbp]
\centering
\caption{Plug-and-Play Robustness on Noisy Backbones. Performance (mIoU \%) evaluated on the PASCAL VOC segmentation task. By upgrading the standard NAF baseline with our multi-VFM relational guidance (NAF w/ DiveUp), the upsampling performance is consistently improved.}\vspace{-2mm}
\label{tab:supp_plug_play}
\resizebox{0.75\textwidth}{!}{%
\begin{tabular}{lcc|c}
\Xhline{1.2pt}
\multirow{2}{*}{\textbf{Target VFM}} & \multicolumn{2}{c|}{\textbf{Semantic Segmentation (mIoU \%)}} & \multirow{2}{*}{\textbf{Absolute Gain}} \\ \cmidrule(lr){2-3}
 & \textbf{NAF (Baseline)} & \cellcolor[HTML]{E7E7FF}\textbf{NAF w/ DiveUp} & \\ 
\hline\hline
PE-Core-S     & 74.37 & \cellcolor[HTML]{E7E7FF}\textbf{75.91} & \textcolor{blue}{+1.54} \\
DFN-CLIP-B & 78.04 & \cellcolor[HTML]{E7E7FF}\textbf{78.78} & \textcolor{blue}{+0.74} \\
MaskCLIP-B      & 73.99 & \cellcolor[HTML]{E7E7FF}\textbf{75.16} & \textcolor{blue}{+1.17} \\
ClearCLIP-B    & 70.98 & \cellcolor[HTML]{E7E7FF}\textbf{73.70} & \textcolor{blue}{+2.72} \\
\Xhline{1.2pt}
\end{tabular}%
}
\end{table}

\section{Additional Ablation Studies}

\subsection{Impact of Local Window Size for Relational Features}
In DiveUp, the local window size ($w \times w$) determines the spatial receptive field for extracting the intrinsic geometric structures when computing the local self-affinity and the center-of-mass (COM) field. To evaluate its impact, we conduct an ablation study using the SigLIP-B backbone on the PASCAL VOC segmentation task.

As shown in Table~\ref{tab:supp_window_size}, the upsampling performance shows an inverted U-shape trend with respect to the window size. A small window (e.g., $3 \times 3$) degrades results (78.94\% mIoU), as the restricted receptive field fails to capture sufficient contextual information for reliable spatial affinity. Conversely, a large window (e.g., $11 \times 11$) leads to performance degradation (79.29\% mIoU). This occurs because broad windows incorporate distant, semantically irrelevant pixels, which injects noise into the self-affinity distribution and smooths out the boundary representations. The optimal balance is achieved at $w = 7$, which captures local geometric structures while filtering out distant spatial noise. Consequently, we adopt $w = 7$ as the default configuration.

\renewcommand{\arraystretch}{1.25}
\begin{table}[htbp]
\centering
\caption{Ablation on Local Window Size ($w \times w$). Performance evaluated on PASCAL VOC using the SigLIP-B backbone. The upsampling performance peaks at $w=7$, which balances spatial context with the exclusion of distant noise.}\vspace{-2mm}
\label{tab:supp_window_size}
\resizebox{0.55\textwidth}{!}{%
\begin{tabular}{c|cc}
\Xhline{1.2pt}
\multirow{2}{*}{\textbf{Window Size ($w$)}} & \multicolumn{2}{c}{\textbf{Semantic Segmentation}} \\ \cmidrule(lr){2-3}
 & \textbf{mIoU (\%)} & \textbf{Acc (\%)} \\ 
\hline\hline
$3 \times 3$  & 78.94 & 94.70 \\
$5 \times 5$  & 79.55 & 94.88 \\
\rowcolor[HTML]{E7E7FF}
\textbf{$7 \times 7$ (Default)} & \textbf{79.60} & \textbf{94.90} \\
$9 \times 9$  & 79.43 & 94.84 \\
$11 \times 11$& 79.29 & 94.80 \\
\Xhline{1.2pt}
\end{tabular}%
}
\end{table}

\section{Deep Dive into Geometric Representations}

\subsection{Visualizing Relational Guidance: Entropy and COM Field}

To further investigate the performance improvements on noisy backbones reported in Section S1.2, we visualize the intermediate relational representations (the local entropy map and the COM field) before and after applying the multi-VFM relational guidance of DiveUp.

\begin{figure*}[htbp]
    \centering
    \includegraphics[trim={0cm 0cm 0cm 0cm}, clip, width=\textwidth]{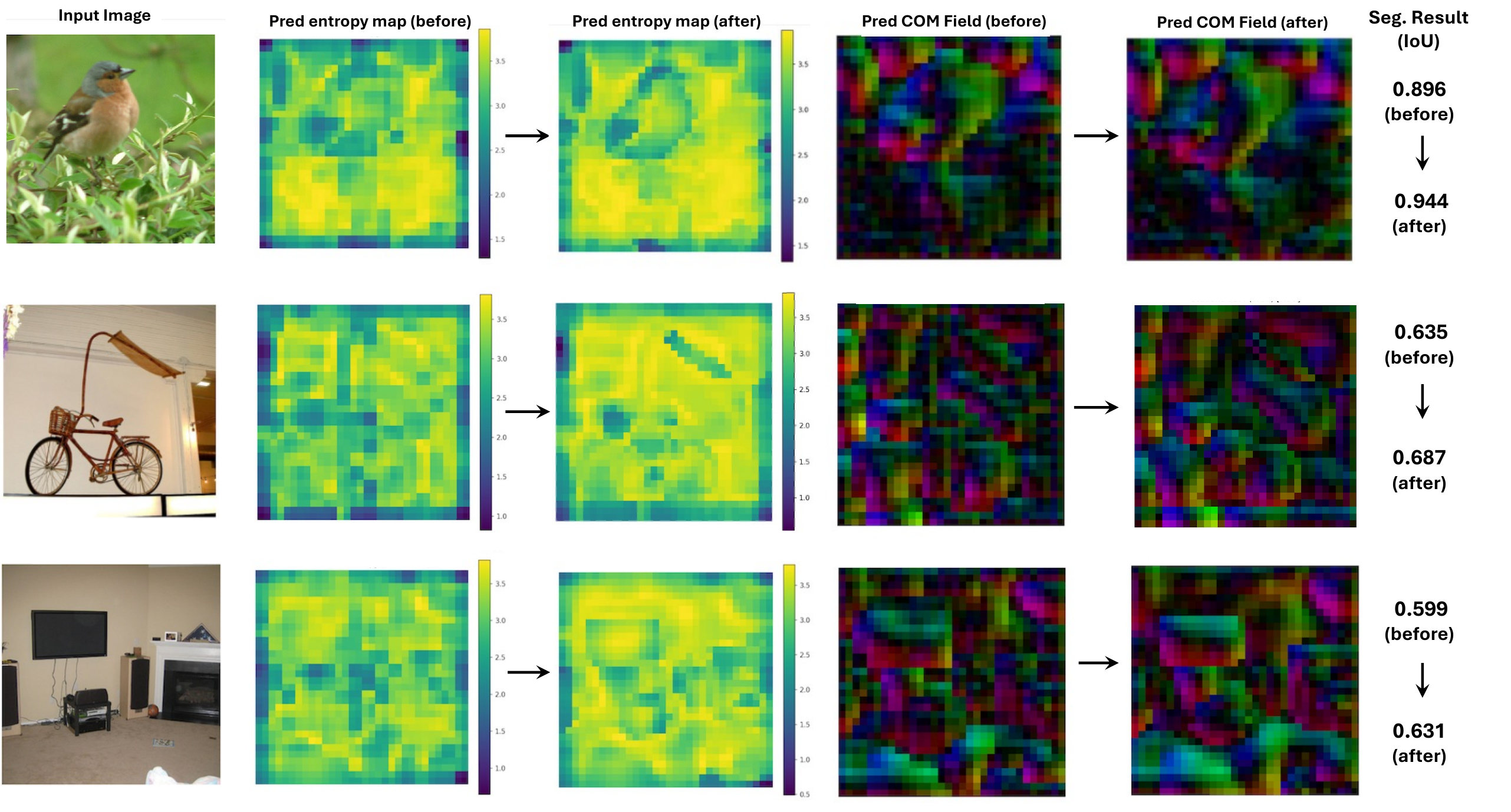} 
    \vspace{-15pt}
    \caption{\textbf{Visualization of Relational Representations (Before vs. After DiveUp).} We visualize the intermediate local entropy maps and COM fields generated by a noisy backbone (ClearClip-B), compared with those generated after integrating DiveUp's multi-VFM relational guidance. DiveUp suppresses background interference and spatial uncertainty, achieving more localized geometric structures. This improved location alignment corresponds to the quality of the final segmentation results (see IoU gains).}
    \label{fig:supp_com_vis}
\end{figure*}

Figure~\ref{fig:supp_com_vis} illustrates this comparison. 
\textbf{Before DiveUp (Baseline):} The native features of the target VFM exhibit high spatial ambiguity. In the baseline entropy maps, severe location misalignment is evident through the fuzzy, intermingled green and yellow regions. This mixing blurs the boundaries between semantic interiors and the background, failing to isolate objects cleanly. As a result, the corresponding COM fields contain misaligned directional vectors and background interference.

\textbf{After DiveUp:} When optimized with our \textit{Consensus COM Field}, the intrinsic geometric structures become better location-aligned. The predicted entropy maps effectively suppress internal and background noise, producing clearer object contours. Similarly, the predicted COM fields (visualized via HSV maps) align more accurately with actual object boundaries and show fewer background artifacts. 

These refined geometric representations correct the location misalignments inherent in the source VFM. This structural improvement contributes to the enhanced downstream segmentation performance, which is consistent with the absolute IoU gains observed in the comparisons.

\section{Comprehensive Qualitative Results}

We provide qualitative comparisons against baseline methods on semantic segmentation and dense depth estimation tasks to support the quantitative results. 

\subsection{High-Resolution Semantic Segmentation}
\begin{figure*}[htbp]
    \centering
    \includegraphics[width=\textwidth]{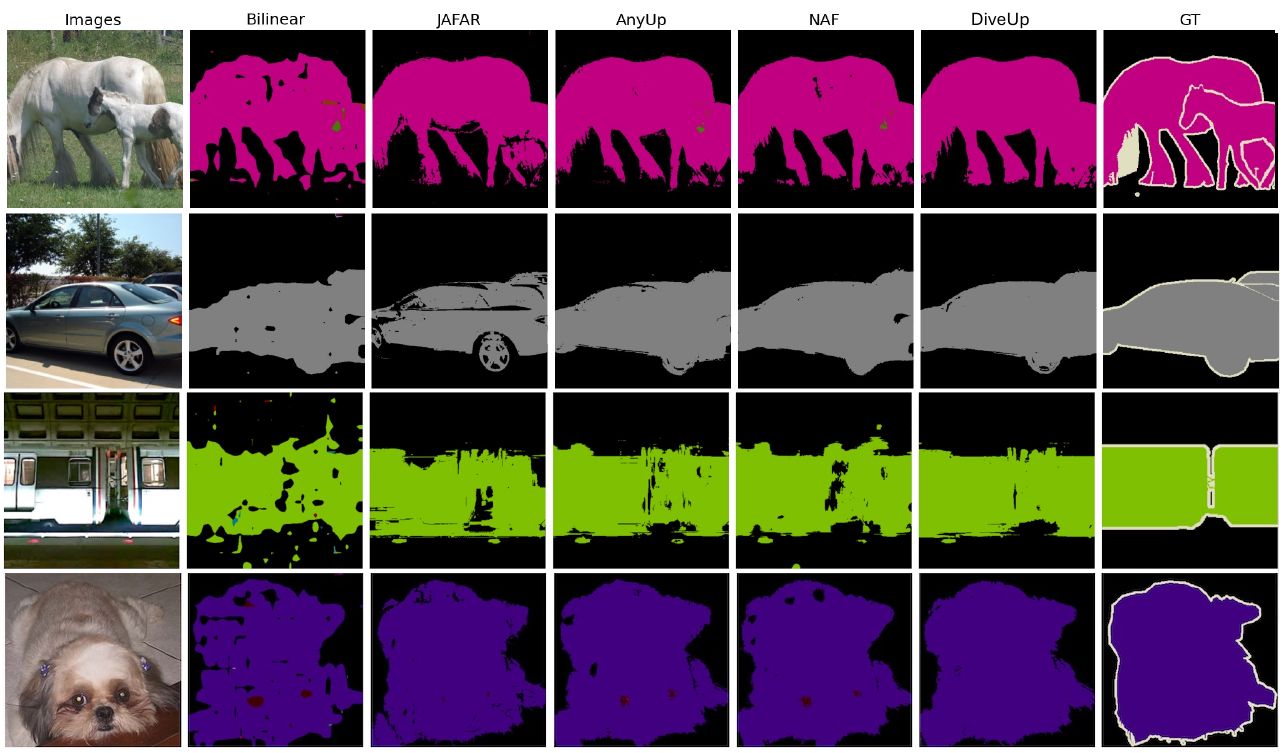} 
    \vspace{-15pt}
    \caption{\textbf{Qualitative Comparison for Semantic Segmentation.} Results are produced using the SigLIP-B backbone on the PASCAL VOC dataset. Compared to other VFM-agnostic and VFM-specific upsamplers, DiveUp eliminates semantic bleeding and generates sharper object boundaries, reflecting the intrinsic geometric structures.}
    \label{fig:supp_seg_vis}
\end{figure*}

In Figure~\ref{fig:supp_seg_vis}, we visualize the semantic segmentation results generated using features from the SigLIP-B backbone. As a vision-language model, SigLIP exhibits location misalignment and high-norm artifacts. 

As observed in the visual comparisons, baseline upsamplers (including JAFAR, AnyUp, and NAF) fail to fully correct these artifacts, resulting in blurry object boundaries and semantics leaking into the background. In contrast, DiveUp leverages the \textit{Consensus COM Field} to confine the semantic representations within the object boundaries. This results in sharper segmentation masks that align with the Ground Truth (GT).

\subsection{Dense Depth Estimation}
\begin{figure*}[htbp]
    \centering
    \includegraphics[width=\textwidth]{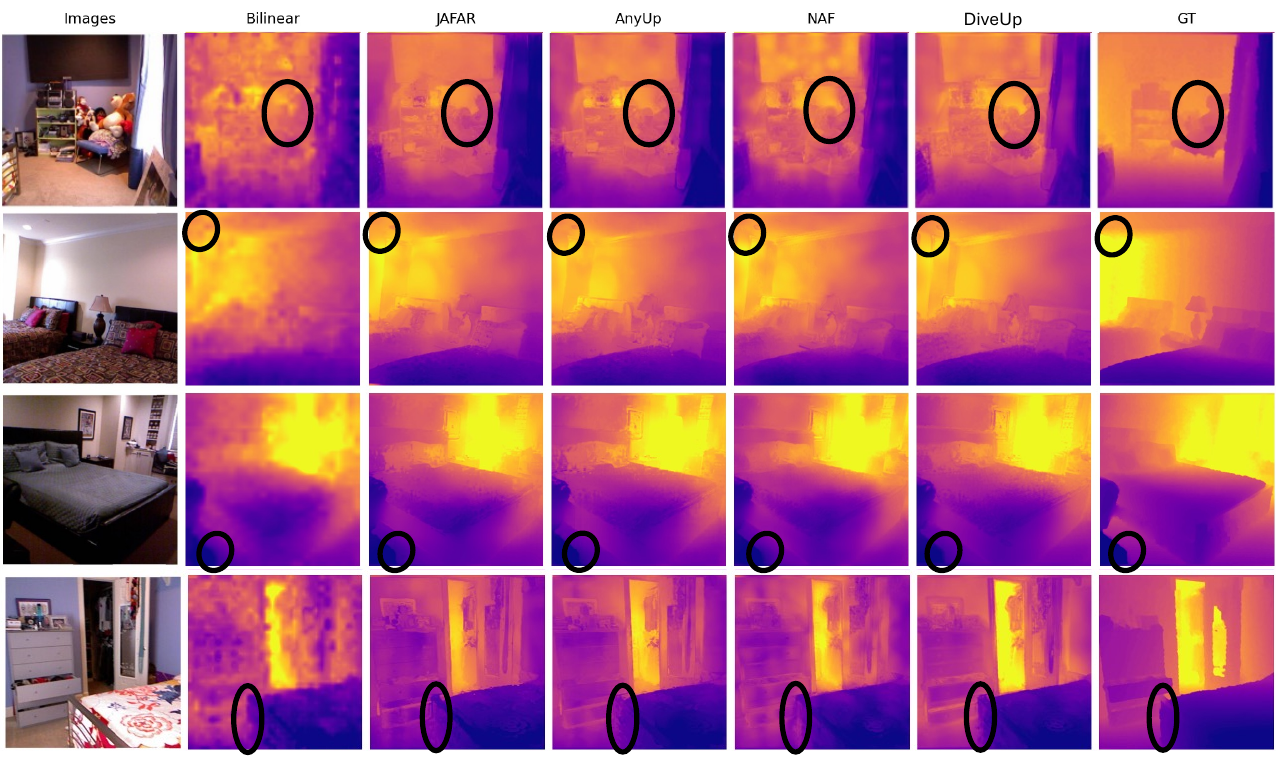} 
    \vspace{-15pt}
    \caption{\textbf{Qualitative Comparison for Dense Depth Estimation.} Results are produced using the DINOv2-S backbone on the NYUv2 dataset. DiveUp preserves depth discontinuities and complex object boundaries.}
    \label{fig:supp_depth_vis}
\end{figure*}

In Figure~\ref{fig:supp_depth_vis}, we compare the qualitative results for dense depth estimation using the DINOv2-S backbone. While DINOv2 natively provides strong patch-level representations, standard upsampling techniques often introduce jagged edges or blocky artifacts at boundaries. By aligning the upsampling process with our \textit{Consensus COM Field}, DiveUp preserves these structural nuances. 

\section{Detailed Architecture and Evaluation Protocols}

\subsection{Upsampler Architecture}
Our upsampler $\mathcal{U}_\theta$ adopts a lightweight, VFM-agnostic spatial interpolation design inspired by NAF~\cite{chambon2025nafzeroshotfeatureupsampling}. It utilizes a cross-scale neighborhood attention mechanism to fuse high-resolution spatial structures with low-resolution VFM semantics.

The complete forward pass is formalized in Algorithm~\ref{alg:upsampler_code}. Specifically, we extract position-aware spatial guidance from the high-resolution image $I_{hr}$ to formulate both the Queries ($Q_{hr}$) and the pooled Keys ($K_{lr}$). To preserve the original semantic space of the target VFM, the Values ($V_{lr}$) inherit the raw features $F_{lr}$ directly without any linear projection. A localized attention operation then aligns these original VFM semantics to the high-resolution spatial queries, yielding the final output $\hat{F}_{hr}$.

\renewcommand{\algorithmicrequire}{\textbf{Input:}}
\renewcommand{\algorithmicensure}{\textbf{Output:}}
\renewcommand{\algorithmiccomment}[1]{\hfill\textcolor{gray}{// #1}}

\begin{algorithm}[htbp] 
\caption{Architecture of the Upsampler $\mathcal{U}_\theta$}
\label{alg:upsampler_code}
\begin{algorithmic}[1]
\Require Low-resolution VFM features $F_{lr} \in \mathbb{R}^{h \times w \times C}$, High-resolution image $I_{hr} \in \mathbb{R}^{H \times W \times 3}$, Window size $w$

\State \textcolor{gray}{// 1. Spatial Guidance Extraction}
\State Extract guidance features: $G = \text{ImageEncoder}(I_{hr})$ \Comment{$G \in \mathbb{R}^{H \times W \times d}$}
\State Apply position embeddings: $G_{RoPE} = \text{Apply2DRoPE}(G)$ 

\State \textcolor{gray}{// 2. Query, Key, and Value Formulation}
\State Formulate queries: $Q_{hr} = G_{RoPE}$ \Comment{High-resolution spatial queries}
\State Formulate keys: $K_{lr} = \text{AvgPool2D}(G_{RoPE})$ \Comment{Low-resolution spatial keys}
\State Formulate values: $V_{lr} = F_{lr}$ \Comment{Raw low-resolution VFM semantics}

\State \textcolor{gray}{// 3. Cross-scale Neighborhood Attention}
\State Predict features: $\hat{F}_{hr} = \text{NeighborhoodAttention}(Q_{hr}, K_{lr}, V_{lr}, w)$ \Comment{Local spatial aggregation}

\State \textbf{return} $\hat{F}_{hr}$
\end{algorithmic}
\end{algorithm}

\subsection{Linear Probing Evaluation Protocols}
To evaluate the representation quality of the upsampled features $\hat{F}_{hr}$, we follow the linear probing protocol used in prior work ~\cite{couairon2025jafar, chambon2025nafzeroshotfeatureupsampling}. 

During the evaluation phase, both the VFM backbone $\mathcal{B}$ and the trained upsampler $\mathcal{U}_\theta$ are frozen. The probing head consists of a single $1 \times 1$ convolutional layer without any non-linear activation functions or multi-layer decoders. The model relies entirely on a raw, pixel-wise linear objective.

The training pipelines for Semantic Segmentation and Dense Depth Estimation are detailed in Algorithm~\ref{alg:probe_seg} and Algorithm~\ref{alg:probe_depth}, respectively. Specifically for depth estimation, as existing benchmarks lack standardized probing implementations on NYUv2, we adapt the protocol from JAFAR~\cite{couairon2025jafar}. However, to ensure a fair evaluation of the local geometric structures, we discard the global \texttt{[CLS]} token from the VFM. The linear regression is applied exclusively to the dense spatial features. 

\begin{algorithm}[!htbp]
\caption{Linear Probing for Semantic Segmentation}
\label{alg:probe_seg}
\begin{algorithmic}[1]
\Require High-resolution image $I \in \mathbb{R}^{H \times W \times 3}$, Ground Truth Mask $M_{gt} \in \mathbb{R}^{H \times W}$
\Require Frozen VFM Backbone $\mathcal{B}$, Frozen Upsampler $\mathcal{U}_\theta$
\Require Trainable $1 \times 1$ Conv Weight $W_{seg} \in \mathbb{R}^{N_{cls} \times C \times 1 \times 1}$

\State \textcolor{gray}{// 1. Feature Extraction (No Gradients)}
\State Extract low-res semantics: $F_{lr} = \mathcal{B}(I)$ \Comment{Frozen backbone forward}
\State Predict high-res features: $\hat{F}_{hr} = \mathcal{U}_\theta(I, F_{lr})$ \Comment{Frozen upsampler forward}
\State Detach from computation graph: $\hat{F}_{hr} = \text{StopGradient}(\hat{F}_{hr})$

\State \textcolor{gray}{// 2. Linear Classification}
\State Predict pixel-wise logits: $S = W_{seg} * \hat{F}_{hr}$ \Comment{$S \in \mathbb{R}^{N_{cls} \times H \times W}$}
\State Compute objective: $\mathcal{L}_{CE} = \text{CrossEntropy}(S, M_{gt})$ 

\State \textcolor{gray}{// 3. Optimization}
\State \textbf{return} $\nabla_{W_{seg}} \mathcal{L}_{CE}$ \Comment{Update ONLY the linear classifier}
\end{algorithmic}
\end{algorithm}

\vspace{-15pt}
\begin{algorithm}[htbp]
\caption{Linear Probing for Dense Depth Estimation}
\label{alg:probe_depth}
\begin{algorithmic}[1]
\Require High-resolution image $I \in \mathbb{R}^{H \times W \times 3}$, Ground Truth Depth $D_{gt} \in \mathbb{R}^{H \times W}$
\Require Frozen VFM Backbone $\mathcal{B}$, Frozen Upsampler $\mathcal{U}_\theta$
\Require Trainable $1 \times 1$ Conv Weight $W_{dep} \in \mathbb{R}^{256 \times 2C \times 1 \times 1}$ \Comment{256 depth bins}

\State \textcolor{gray}{// 1. Feature Extraction (No Gradients)}
\State Extract dense semantics: $F_{lr} = \mathcal{B}(I)$ 
\State Predict high-res features: $\hat{F}_{hr} = \mathcal{U}_\theta(I, F_{lr})$ \Comment{Frozen upsampler forward}
\State Detach from computation graph: $\hat{F}_{hr} = \text{StopGradient}(\hat{F}_{hr})$

\State \textcolor{gray}{// 2. Bin-based Depth Classification}
\State Concatenate features: $F_{in} = \text{Concat}(\hat{F}_{hr}, \hat{F}_{hr})$ \Comment{$F_{in} \in \mathbb{R}^{2C \times H \times W}$}
\State Predict bin logits: $S = W_{dep} * F_{in}$ \Comment{$S \in \mathbb{R}^{256 \times H \times W}$}
\State Normalize probabilities: $P = \frac{\text{ReLU}(S) + 0.1}{\sum \left( \text{ReLU}(S) + 0.1 \right)}$ \Comment{Soft depth assignment}
\State Expected depth: $D = \sum_{k=1}^{256} P_k \cdot \text{bin}_k$ \Comment{Linear combination of bins}

\State \textcolor{gray}{// 3. Optimization}
\State Compute objective: $\mathcal{L}_{depth} = \text{SigLoss}(D, D_{gt}) + \text{GradientLoss}(D, D_{gt})$
\State \textbf{return} $\nabla_{W_{dep}} \mathcal{L}_{depth}$ \Comment{Update ONLY the linear regressor}
\end{algorithmic}
\end{algorithm}

\FloatBarrier 

\end{document}